%% file: combined.tex
\documentclass[10pt,twocolumn,letterpaper]{article}

\PassOptionsToPackage{table}{xcolor}
\usepackage[pagenumbers]{wacv}

\input{preamble}

\usepackage{fontawesome5} 

\definecolor{wacvblue}{rgb}{0.21,0.49,0.74}
\usepackage[pagebackref,breaklinks,colorlinks,allcolors=wacvblue]{hyperref}

\newcommand{\suppref}[2]{\hyperref[#1]{\ref*{#1}}}

\def\confName{WACV}
\def\confYear{2027}

\title{Points, Not Pixels: Keypoints, Masks and Oblique Geometry
for \\Floating-Roof Tank Fill Estimation from Overhead Imagery}
\author{Mohd Shahid Kaleem\thanks{\texttt{mohd.kaleem@students.iiit.ac.in}} \quad Mohd Hozaifa Khan\thanks{\texttt{mohd.hozaifa@research.iiit.ac.in}} \quad Maaz Khan\thanks{\texttt{maaz.khan@research.iiit.ac.in}} \\
International Institute of Information Technology, Hyderabad 500 032, India\\
{\footnotesize\scalebox{0.8}{\faGithub}\: \url{github.com/Shahid0208/pt-not-pix-fill-tank}} \\
{\footnotesize All authors contributed equally to this work.}}

\usepackage{docmute}

\begin{document}

\let\combinedbibliography\bibliography
\let\combinedbibliographystyle\bibliographystyle
\renewcommand{\bibliography}[1]{}
\renewcommand{\bibliographystyle}[1]{}

\input{geocv}

\clearpage
{
    \small
    \combinedbibliographystyle{ieeenat_fullname}
    \combinedbibliography{main}
}

\clearpage
\twocolumn[{\centering\Large\bfseries Supplementary Material\par\vspace{1em}}]

\def\maketitle{}

\input{supplement}

\end{document}

%% file: preamble.tex
\usepackage{amsmath,amssymb}

%% file: geocv.tex
\maketitle

\begin{abstract}
The floating roof of an oil storage tank rests on the stored liquid, so its depth below the rim indicates how full the tank is. In optical overhead imagery, this depth appears as a narrow shadow cast by the rim onto the roof. Measuring the shadow is difficult, and nadir-based height conversions err in oblique views. We predict sub-pixel keypoints on the rim and shadow boundaries and use an affine shadow model to estimate roof depth. We introduce SynTank-25k, with 24,987 physically based renders and exact roof, shadow and camera geometry. On the synthetic test set, the keypoint pipeline has $27\%$ lower error in estimated roof state under the affine estimator than a segmentation pipeline matched in backbone, input size and crop. The gap is largest for shadows narrower than $6\text{ px}$ in the synthetic images, where ground-truth masks converted by region fitting also have higher error than learned keypoints. On wide shadows, the observed difference is small. At $60^\circ$--$65^\circ$ camera elevation, the affine estimator has one twelfth of the nadir formula's error using the same predicted keypoints. On NAIP, a detector trained only on SynTank-25k finds $93\%$ of floating-roof test tanks but produces more false positives than the real-trained detector. Of 6,558 tank crops, about half pass geometric and estimator-agreement checks. Data and code are available at \href{https://doi.org/10.5281/zenodo.24395104}{Zenodo} and \href{https://github.com/Shahid0208/pt-not-pix-fill-tank}{GitHub}.
\end{abstract}

\section{Introduction}
\label{sec:intro}

The roof of an external floating-roof tank rests on the stored liquid, so its depth below the top rim ($\Delta_R$) indicates the liquid level and, together with tank dimensions, supports estimates of stored volume~\cite{villamillopez2021monitoring,wang2019estimating}. Measuring this depth from overhead imagery would allow tanks to be monitored remotely and at scale. Spaceborne synthetic aperture radar (SAR) can track roof depth through wall--roof double-bounce scattering~\cite{villamillopez2021monitoring,ma2022structural,chen2026engineering}, while optical aerial and satellite imagery offers higher spatial resolution and frequent revisits. In optical imagery, however, roof depth is visible only indirectly, through the crescent shadow that the rim casts onto the recessed roof~\cite{wang2019estimating}.

This shadow is small: in the public NAIP storage-tank dataset ($0.6\text{ m/pixel}$)~\cite{robinson2024remotely}, we measure a median floating-roof tank diameter of $56\text{ px}$, and the interior shadow can be only a few pixels wide. A common optical approach segments the shadow, measures its width, and converts the width to height with solar-elevation trigonometry~\cite{irvin1989methods,liasis2016satellite,wang2019estimating}. Depth estimates depend on how the shadow boundary is represented and how its displacement is converted to height. Binary masks place the boundary on a pixel grid, while nadir-based conversions omit camera parallax. Commercial overhead imagery is often off-nadir~\cite{christie2020learning}, which can degrade overhead perception models~\cite{weir2019spacenet}.

\begin{figure*}[t]
\centering
\includegraphics[width=0.85\textwidth,trim=6bp 0bp 6bp 0bp,clip]{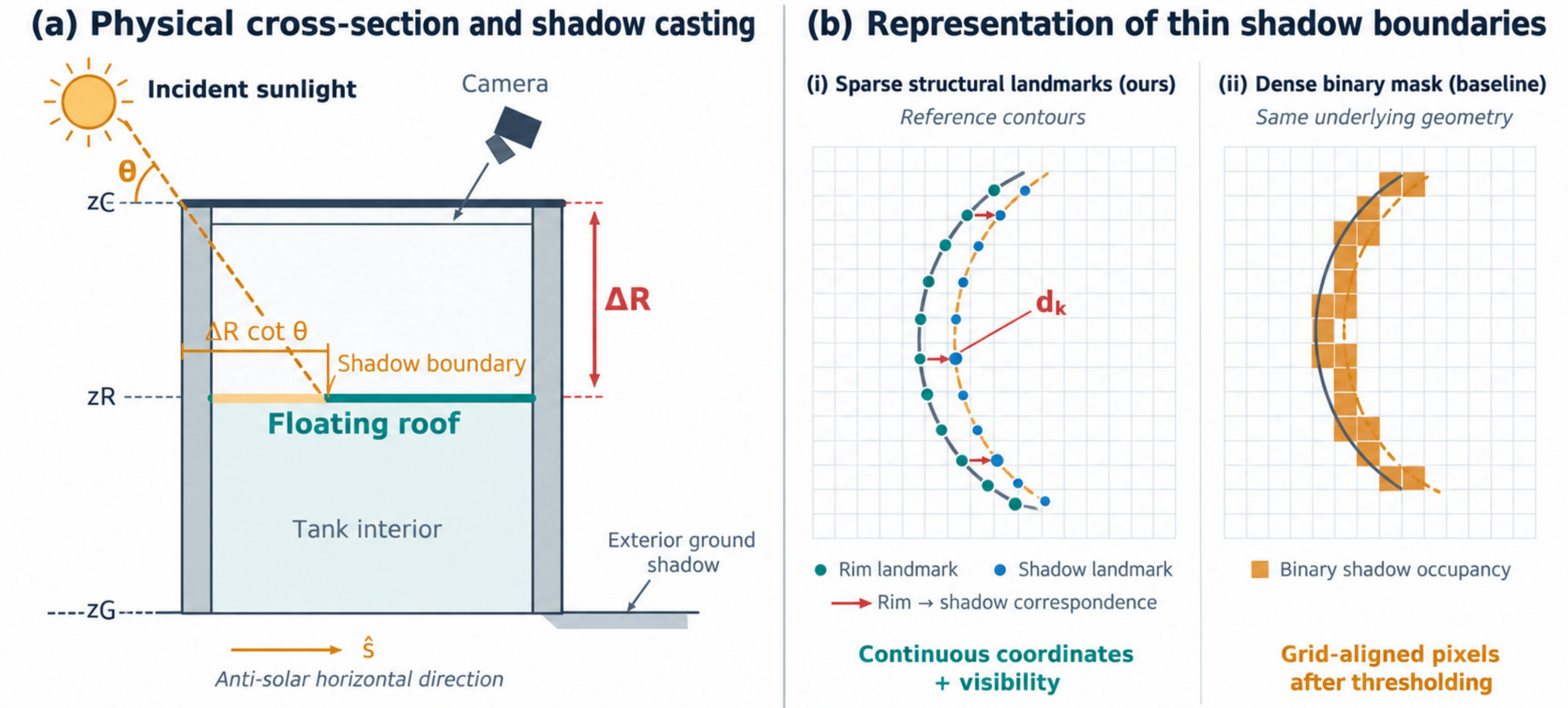}
\caption{\textbf{Shadow geometry and boundary representations for floating-roof tank fill estimation.} (a) Roof depth \(\Delta_R=z_C-z_R\) produces a horizontal shadow displacement \(\Delta_R\cot\theta\) in the anti-solar direction, where \(\theta\) is solar elevation. (b) Two representations of the thin shadow boundary: sub-pixel keypoints with rim-to-shadow correspondences \(\mathbf d_k\), and a binary mask on the pixel grid.}
\label{fig:representations}
\end{figure*}

Under a local affine camera, the rim-to-shadow displacement is proportional to roof depth, with the proportionality set by sun and camera geometry (Fig.~\ref{fig:representations} and Eq.~\eqref{eq:shadow_vec}). We predict sub-pixel keypoints on the boundaries and compare this approach with a pipeline that segments the shadow and fits the resulting mask. Because NAIP has bounding boxes but no fill or roof-height labels~\cite{robinson2024remotely}, we measure fill accuracy on physically based renders with exact geometry. We address three questions:
\begin{enumerate}[labelindent=0pt,labelwidth=2.2em,labelsep=0.5em,leftmargin=2.7em,align=left]
    \item[\textbf{RQ1}] How do keypoint and segmentation pipelines compare in fill accuracy across shadow widths?
    \item[\textbf{RQ2}] How much does the nadir assumption cost under oblique views?
    \item[\textbf{RQ3}] When does the keypoint pipeline fail, and where does the segmentation pipeline lose accuracy?
\end{enumerate}

\paragraph{Contributions.}
\begin{itemize}
    \item We introduce \textbf{SynTank-25k}, with 24,987 validated ray-traced renders in 18,608 scene groups, geometry and acquisition labels, and a sensor model tuned against real NAIP imagery. We evaluate sim-to-real detector transfer and the consistency of keypoint-based estimates on NAIP imagery (Secs.~\ref{sec:datasets} and~\ref{sec:detection}).
    \item We present an affine shadow estimator that accounts for camera parallax and measure the cost of the nadir assumption across camera elevations (Secs.~\ref{sec:model} and~\ref{sec:rq2}).
    \item We compare keypoint and segmentation pipelines with matched backbone, input size and crop. The keypoint advantage is largest on thin shadows (Secs.~\ref{sec:rq1} and~\ref{sec:rq3}).
\end{itemize}


\section{Related Work}
\label{sec:related}

\paragraph{Shadow-based tank measurement.}
Classical overhead mensuration extracts cast shadows, measures their length along the solar direction, and converts length to height with the solar elevation~\cite{irvin1989methods,liasis2016satellite}. Wang et al.~\cite{wang2019estimating} estimated exterior volume from sub-pixel shadow length for 24 fixed-roof tanks of one height in a single Gaofen-2 scene. Exterior shell height is static, whereas floating-roof inventory is governed by the interior roof depth. Li et al.~\cite{li2026reserve} derived coordinate transformations for arc-shaped shadows on floating-roof tanks under oblique views and validated them on a single tank using manual curve fitting. A patented method matches per-fill templates to the image~\cite{patent10192323}. We model off-nadir parallax as linear in roof depth and evaluate on thousands of tanks.


\paragraph{SAR-based monitoring.}
SAR has been widely used for floating-roof tanks because it is weather independent~\cite{hammer2017simulation,villamillopez2021monitoring,ma2021end,ma2022structural,10329403,chen2026engineering}. High-resolution SAR represents tanks with a few strong returns, which support sparse-point measurements: Tsai et al.~\cite{10329403} use five scattering points per tank to estimate tank and roof heights, while Ma et al.~\cite{ma2022structural} estimate storage from structural projection points. For geometric extraction, Chen et al.~\cite{chen2026engineering} compared a Hough-arc pipeline with U-Net segmentation on TerraSAR-X. In optical imagery, roof depth is carried by a continuous shadow boundary only a few pixels wide, and we compare sparse points and masks for measuring it.

\paragraph{Synthetic overhead imagery.}
RarePlanes~\cite{shermeyer2021rareplanes} showed that synthetic aircraft images reduce the real data needed for overhead detection, and Synthinel-1~\cite{kong2020synthinel} that synthetic overhead imagery improves building segmentation in unseen cities. SynTank-25k provides roof, shadow and acquisition geometry to evaluate fill estimation.

\paragraph{Keypoints versus masks.}
Pixel masks are the default output for overhead structures, but mask quality and geometric accuracy can diverge. Mask IoU weighs all pixels equally, so on large regions it is insensitive to boundary errors~\cite{cheng2021boundary,kirillov2019pointrend}, and networks that predict labels on a coarse regular grid undersample object boundaries and blur contours~\cite{kirillov2019pointrend}. For building outlines, PolyWorld~\cite{9880425} detects polygon vertices directly instead of vectorizing a segmentation mask and outperforms mask-based polygonization. Heatmap keypoints~\cite{xiao2018simple} carry their own quantization error when decoded from a reduced-resolution heatmap, which sub-pixel encoding and decoding are designed to reduce~\cite{9157744}. We compare the two representations on a measurement task, where boundary errors affect the recovered fill through the shadow geometry.
\section{Shadow Geometry and Estimators}
\label{sec:model}

\paragraph{Quantities.}
Let $z_C$, $z_R$, $z_G$ and $z_F$ denote the elevations of the rim crest, the floating roof, the exterior ground, and the tank floor. With roof depth $\Delta_R = z_C - z_R$ and wall height above ground $H_G = z_C - z_G$, the ground-referenced normalized roof state is
\begin{equation}
    \phi_R = \frac{z_R - z_G}{z_C - z_G} = 1 - \frac{\Delta_R}{H_G} \in [0, 1].
    \label{eq:phi_r}
\end{equation}
We measure roof-state error in $\phi_R$ and report its mean absolute error (MAE). The liquid fill fraction is $(z_R - z_F)/(z_C - z_F)$ and equals $\phi_R$ only when $z_F = z_G$; in SynTank-25k the deck rests on its legs at $z_F = 0.2\text{ m}$ above the ground, so the two differ slightly. We calculate volume as a geometric cylinder, $V = \pi R^2 (z_R - z_F)$, and report it as a fraction of capacity because relative errors diverge for nearly empty tanks.

\paragraph{Affine imaging model.}
Over a single tank footprint, a satellite or aerial sensor is well approximated by a local affine projection~\cite{christie2020learning}
\begin{equation}
    y = A x + b z + c,
    \label{eq:affine_cam}
\end{equation}
where $x \in \mathbb{R}^2$ is the horizontal ground coordinate, $z$ the elevation, $y$ the image coordinate, $A$ the local ground-to-image Jacobian, and $b$ the off-nadir parallax vector (pixels per metre of elevation). For solar elevation $\theta$ and horizontal anti-solar direction $\hat{s}$, the rim's shadow falls $\Delta_R \cot\theta\,\hat{s}$ away on the roof plane, $\Delta_R$ below the rim. Projecting both points gives the image displacement of the interior shadow boundary relative to the rim:
\begin{equation}
    \delta_1 = \Delta_R\, a_s, \qquad a_s = \cot\theta\, A\hat{s} - b.
    \label{eq:shadow_vec}
\end{equation}
The exterior ground shadow is displaced by $\delta_3 = H_G\, a_s$. All depth information therefore lies in one boundary displacement along $a_s$.

\paragraph{Estimators.}
We use three estimators with increasing metadata requirements: a shadow-displacement ratio (M1), a nadir estimator (M2), and an affine estimator (M3). M1 uses no acquisition metadata and tests whether a representation recovers both shadow boundaries. M2 uses solar elevation and ground sample distance; M3 also uses the camera model. Applying M2 and M3 to the same predicted keypoints compares the errors of the nadir and affine conversions (Sec.~\ref{sec:rq2}).
\begin{itemize}
    \item \textbf{Shadow-displacement ratio (M1):} $\hat{\phi}_R = 1 - \|\delta_1\| / \|\delta_3\|$. Ratios of interior to exterior measurements are established practice, using crescent-shadow areas or rim-to-roof and rim-to-base lengths~\cite{patent8842874}; we also evaluate the area form (Sec.~\ref{sec:estimators}). Under Eq.~\eqref{eq:shadow_vec} the shadow-displacement form is exact because $a_s$ cancels. It assumes only a nominal shell thickness ($0.35\text{ m}$ in our synthetic images) to locate the inner rim, and it requires an unoccluded exterior shadow.
    \item \textbf{Nadir estimator (M2):} $\hat{\Delta}_R = \|\delta_1\|\, g \tan\theta$ with ground sample distance $g$, the standard shadow-to-height conversion~\cite{irvin1989methods,wang2019estimating}. The nadir estimator assumes $b = 0$, so the shadow displacement is attributed entirely to the sun.
         \item \textbf{Affine estimator (M3):} Each displacement $d_k$ runs from an inner-rim point to its corresponding interior-shadow boundary point. Eq.~\eqref{eq:shadow_vec} is linear in $\Delta_R$, so a one-dimensional robust fit over $K$ such displacements estimates the roof depth,
    \begin{equation}
        \hat{\Delta}_R = \arg\min_{\Delta_R} \sum_{k=1}^{K} \rho\big(\|d_k - \Delta_R\, a_s\|\big),
        \label{eq:m3_huber}
    \end{equation}
    with Huber loss $\rho$ ($1\text{ px}$), minimized over roof elevations on a $0.005\text{ m}$ grid with parabolic refinement. The affine estimator accounts for parallax $b$ and uses the acquisition metadata ($A$, $b$, $\theta$, $\hat{s}$) and the wall height $H_G$ to convert $\hat{\Delta}_R$ to $\hat{\phi}_R$ through Eq.~\eqref{eq:phi_r}.

\end{itemize}

\section{SynTank-25k}
\label{sec:datasets}

\paragraph{Why a synthetic benchmark.}
The datasets in Table~\ref{tab:compare} label tank locations, instances, or shell attributes. Fill evaluation requires roof-position labels matched to image acquisition time, such as operator gauge records. SynTank-25k supplies geometric truth and varies sun and view across scenes. In real archives, sun and view co-vary with site, season, and sensor~\cite{weir2019spacenet}. Following RarePlanes~\cite{shermeyer2021rareplanes}, we pair the synthetic set with real NAIP imagery to measure transfer (Sec.~\ref{sec:detection}).

\begin{table}[t]
\centering
\caption{Storage-tank datasets and their per-tank labels. Of the listed datasets, only SynTank-25k includes roof-position and fill labels. OST and HSC are described in Robinson et al.~\cite{robinson2024remotely}.}
\label{tab:compare}
\resizebox{\columnwidth}{!}{%
\begin{tabular}{@{}llll@{}}
\toprule
\textbf{Dataset} & \textbf{Imagery} & \textbf{Size} & \textbf{Per-tank labels} \\
\midrule
NAIP AST~\cite{robinson2024remotely} & Aerial, $0.6\text{ m}$ & 27,639 chips, $>$130k tanks & Box, class, diameter, date \\
DOTA~\cite{xia2018dota} & Aerial, mixed & $>$10,000 storage tanks & Oriented box \\
OST & Google Earth & 1,595 images & Floating-roof tank instances \\
HSC & Inventory & $>$4,500 tanks & Location, diameter, height, contents \\
\midrule
\textbf{SynTank-25k} & Synthetic, $0.5$--$0.7\text{ m}$ & 24,987 synthetic images & 3D geometry, roof depth, fill, \\
 & & & sun and camera, boundaries, masks \\
\bottomrule
\end{tabular}%
}
\end{table}

\paragraph{Tanks and scenes.}
Each synthetic image centres one labelled floating-roof tank. Radius is drawn log-uniformly within four size classes; shell height is drawn uniformly and independently of radius, and fill is continuous (Table~\ref{tab:dataset}). The deck ranges from its rest height $z_F$ to $0.1\text{ m}$ below the rim. Tanks vary in seal gap, deck offset and tilt, and flush seal ring. Raised rim structures are excluded because the shadow model places the rim shadow on the deck. Farm layouts include dikes, roads, pipes, buildings, and site-dependent vegetation. Neighboring tanks receive independent heights; those whose shadow or lean would reach the labelled tank are excluded. Figure~\suppref{fig:supp_syntank_examples}{S1} shows examples across sites and roof states.

\paragraph{Sun, view and atmosphere.}
Five sites set latitude, ground colour, and vegetation. Each image uses a real 2025 date and local solar time; the NOAA solar calculator~\cite{noaa_solar} gives the sun position. Sun elevation is balanced over five bands; sites that do not reach a band are supplemented by other sites. Scenes are rendered with Blender Cycles~\cite{blender} through an orthographic camera, matching the affine model in Eq.~\eqref{eq:affine_cam}. Camera elevation is measured from the horizon, with $90^\circ$ denoting nadir. Clear, haze, and thin-cloud atmospheres retain direct sunlight; overcast scenes are excluded. Observability constraints limit the extent of the rim shadow and require part of the roof to remain visible and sunlit. Site-specific acquisition windows and constraint thresholds are given in Appendix~\suppref{supp:dataset}{B}.

\paragraph{Scale and sensor model.}
Half of the synthetic images use NAIP-scale ground sample distance; the other half use log-uniform zoom to include close-ups and wider farm views (Table~\ref{tab:dataset}). A sensor model simulates NAIP-scale image degradation while keeping labels aligned with the final image. Sensor parameters were tuned to measurements from 73 NAIP floating-roof tanks (Cushing, Oklahoma, and three Texas terminals), including lit-to-shadow luminance ratios, shadow colour, noise, and $2.5$--$3.5\text{ px}$ shadow edges. In the zoomed half, objects span more render pixels than sensor pixels; sizes and width strata are reported in render pixels. Appendix~\suppref{supp:dataset}{B} gives the sensor operations and zoom settings.

\paragraph{Balanced design.}
Factors are sampled across balanced bins, with continuous values drawn within each bin. Size class, height (4 bins), and fill (20 bins) are fully crossed; site and sun band are balanced; atmosphere and camera elevation (6 bins) are crossed. Scene groups are shuffled to even out pairwise combinations. Radius and height are nearly uncorrelated (Table~\ref{tab:dataset}). The lit-roof constraint links fill and sun elevation: deeply recessed roofs are not paired with very low sun.

\begin{table}[t]
\centering
\caption{SynTank-25k at a glance. Tank diameter statistics are computed in render pixels over the 4,997 validation and test images.}
\label{tab:dataset}
\resizebox{\columnwidth}{!}{%
\begin{tabular}{@{}ll@{}}
\toprule
\textbf{Property} & \textbf{Value} \\
\midrule
Synthetic images / scene groups & 24,987 / 18,608 \\
Split (train / val / test) & 19,990 / 3,748 / 1,249 synthetic images \\
Radius, height & $6$--$45\text{ m}$ (log-uniform), $9$--$22\text{ m}$ (corr.\ $-0.02$) \\
Fill $\phi_R$ & Continuous, $0.01$--$0.995$ \\
Tank diameter & Median $99\text{ px}$ ($5$--$95\%$: $24$--$299\text{ px}$; $31\% \le 60\text{ px}$) \\
Scale & Half at $0.5$--$0.7\text{ m/px}$, half zoomed; effective ground sample distance $0.5$--$0.7\text{ m}$ \\
Camera elevation & Continuous, $60^\circ$--$90^\circ$, random azimuth \\
Sun elevation & $10^\circ$--$87^\circ$ (NOAA, real 2025 dates and times) \\
Scenes & 6 layouts, 5 sites, 3 atmospheres \\
Per-tank variation & Seal gap, deck offset and tilt, seal ring \\
Truth & 3D world, projected boundaries, metadata, masks, 65 keypoints \\
\bottomrule
\end{tabular}%
}
\end{table}

\paragraph{Ground truth.}
Each synthetic image carries four truth layers: 3D world geometry; projected rim, shadow, and roof boundaries with visibility flags; acquisition metadata ($A$, $b$, $\theta$, $\hat{s}$, ground sample distance, date, and time); and dense masks for tank instances, structural parts, and both shadow regions. Masks are produced before the sensor model and remain sharp. Projected boundaries also define 65 sun-relative keypoints, with visibility flags. Across the dataset, $33.9\%$ of interior-shadow keypoints, $48.8\%$ of exterior-shadow keypoints, and $59.0\%$ of roof-edge keypoints are visible.

\paragraph{Technical validation.}
We validate exported boundaries against analytic geometry, check their alignment with synthetic shadow masks, and test roof geometry and camera projection against the physical model. Appendix~\suppref{supp:validation}{C} gives the checks' sampling rules and thresholds. Of the 25,000 synthetic images, 13 failed validation (12 narrowly, 1 clearly) and were removed. Exported boundaries deviate from the analytic curves by at most $0.001\text{ px}$.

\paragraph{Splits and release.}
The detector, keypoint, and segmentation models share a scene-level split (same site, date, and time), so no scene crosses splits: 19,990 / 3,748 / 1,249 synthetic images in 13,927 / 3,455 / 1,226 scene groups. No byte-identical crops cross splits, and no cross-split crop pairs have thumbnail correlation above $0.99$ (maximum $0.978$). Validation serves checkpoint selection and diagnostics; comparisons are reported on the test split.

\section{Keypoint and segmentation pipelines}
\label{sec:estimators}


\begin{figure}[t]
\centering
\includegraphics[width=\linewidth]{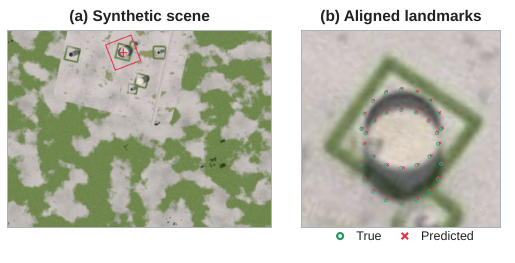}
\caption{Synthetic test example with near-median roof-state error (seed 0). (a) Scene with the selected tank and rotated crop in red. (b) Sun-aligned $320\times320$ context crop with ground-truth and predicted keypoints; the shadow direction points down.}
\label{fig:pipeline}
\end{figure}

\paragraph{Keypoint pipeline.}
A YOLO26s-OBB detector~\cite{yolo26} (DOTA-pretrained) finds tanks with oriented boxes (Fig.~\ref{fig:pipeline}). Each tank is cropped into a $320\times320$ window spanning five tank radii and rotated so that the shadow direction $\hat{s}$ points down; the model sees its central $256\times256$ region, spanning four radii. A ResNet-18~\cite{he2016deep} SimpleBaseline~\cite{xiao2018simple} ($16.5\text{M}$ parameters) predicts 65 heatmaps: the tank centre and 16 sun-relative keypoints on each of four curves (rim crest, interior shadow boundary, exterior ground shadow boundary, roof edge). Points with the same index are constructed from the same sun-relative world azimuth; shadow points are displaced from their casters. Training uses heatmap MSE, visibility cross-entropy, coordinate and displacement smooth-$L_1$ terms, for 35 epochs, with the checkpoint of lowest validation keypoint localization error. Each heatmap is decoded by its argmax followed by distribution-aware Taylor refinement~\cite{9157744}. Keypoints feed the three estimators when their predicted visibility probability exceeds 0.5. M3 abstains when no interior-shadow keypoint is usable; M1 also requires an exterior-shadow keypoint. Training and decoding details are in Appendix~\suppref{supp:training}{D}.

\paragraph{Segmentation baseline.}
A ResNet-18~\cite{he2016deep} U-Net~\cite{ronneberger2015unet} ($13.6\text{M}$ parameters) segments interior and exterior shadows on $256\times256\text{ px}$ crops spanning four tank radii, so its backbone, input size and crop match the keypoint model. It is trained for 35 epochs, with the checkpoint of highest validation inner-shadow IoU. Because the network predicts only shadows, both conversions take the rim from the known tank geometry. For the affine estimator, the model predicts the interior-shadow region at each candidate roof elevation and fits it to the predicted shadow probabilities using fractional pixel coverage. For M1 and M2, the rim-to-shadow displacement is the shift that best aligns the projected rim lip with the predicted shadow. The alignment is undefined when no shift reduces the mismatch by at least $10\%$, a condition that occurs mostly for thin shadows seen obliquely (Appendix~\suppref{supp:geometry}{A}).

\paragraph{Crescent-area reference.}
We also evaluate the crescent-area ratio~\cite{patent8842874} with a solar-elevation correction,
\begin{equation}
    \hat\phi_{\mathrm{area}} = \operatorname{clip}_{[0,1]}\!\left(1-\frac{A_{\mathrm{inner}}}{A_{\mathrm{ground}}\sin\theta}\right),
    \label{eq:area_ratio}
\end{equation}
where the areas count interior and exterior ground-shadow pixels in the predicted class mask and $\theta$ is the known solar elevation. The area estimate uses shadow areas rather than boundary displacements. It is set to zero when $A_{\mathrm{ground}}=0$, so it covers every tank.

\section{Experiments}
\label{sec:experiments}

Both models are trained with four seeds on $4\times$ RTX 2080 Ti GPUs. Unless stated otherwise, we report MAE in $\phi_R$ on the test split as mean $\pm$ standard deviation over the four seeds, on the \emph{common subset}: the test tanks that both pipelines cover under the given estimator. Coverage, the share of test tanks for which a pipeline returns an estimate, is reported alongside. An estimate is a failure when its absolute error exceeds $0.02$, about $0.18$--$0.44\text{ m}$ of roof depth for the $9$--$22\text{ m}$ walls in SynTank-25k.

\subsection{RQ1: Keypoint vs.\ segmentation pipelines}
\label{sec:rq1}

\begin{table}[t]
\centering
\caption{Test roof-state MAE on tanks scored by both pipelines under each estimator, mean $\pm$ standard deviation over four seeds. Coverage is out of the 1,249 test tanks. Ground-truth rows use the corresponding affine readouts for keypoints and masks.}
\label{tab:main}
\resizebox{\columnwidth}{!}{%
\begin{tabular}{@{}lccc@{}}
\toprule
 & \textbf{M3 (affine)} & \textbf{M1 (ratio)$^\dagger$} & \textbf{M2 (nadir)} \\
\midrule
Keypoints & $\mathbf{0.0123 \pm 0.0002}$ & $\mathbf{0.0133 \pm 0.0006}$ & $\mathbf{0.0745 \pm 0.0003}$ \\
\quad coverage & $99\%$ & $75\%$ & $99\%$ \\
Segmentation & $0.0169 \pm 0.0012$ & $0.0205 \pm 0.0005$ & $0.0784 \pm 0.0003$ \\
\quad coverage & $99\%$ & $79\%$ & $82\%$ \\
Common subset (tanks) & 1,234 & 725 & 1,021 \\
\midrule
\multicolumn{4}{@{}l}{\textit{Ground-truth inputs (affine)}} \\
Ground-truth keypoints & $0.0016$ & -- & -- \\
Ground-truth masks & $0.0121$ & -- & -- \\
\bottomrule
\multicolumn{4}{@{}l}{\footnotesize $^\dagger$M1 (displacement ratio) needs an unoccluded exterior shadow.}
\end{tabular}%
}
\end{table}

Across all estimators and seeds, the keypoint pipeline has lower roof-state error than the segmentation pipeline (Table~\ref{tab:main}). For the affine estimator, the error is $27\%$ lower (per-seed results and coverage in Tables~\suppref{tab:supp_seed_results}{S3}--\suppref{tab:supp_own}{S4}). The segmentation network reaches test IoUs of $0.915$ for the interior shadow and $0.935$ for the ground shadow. Median keypoint localization error is $0.72\text{ px}$ for rim and shadow points. IoU weights every pixel equally~\cite{cheng2021boundary}, whereas roof-state error depends on boundary displacement. With ground-truth inputs, keypoints give $0.0016$ error and masks give $0.0121$ through region fitting. The displacement fit adds little error to ground-truth keypoints. With region fitting, the ground-truth mask error is close to the learned keypoint error (Appendix~\suppref{supp:true}{F}).

For segmentation, M2 has lower coverage because the shift search sometimes fails to reduce the mismatch sufficiently. Most abstentions occur on oblique views of thin shadows (median camera elevation $68^\circ$, median width $5\text{ px}$). On these tanks, the keypoint pipeline's M2 error is $0.18$.

For keypoint seed 0, the roof-height MAE is $0.19\text{ m}$ and the volume error $1.3\%$ of capacity on its 1,236 covered test tanks. The crescent-area estimate (Eq.~\eqref{eq:area_ratio}) from the segmentation masks gives $0.230$ on all $1,249$ test tanks, an order of magnitude above the displacement-based estimators.

\begin{table}[t]
\centering
\caption{Test roof-state MAE using the affine estimator by true interior-shadow width (render pixels) on the common subset, mean over four seeds (mean tanks per seed, rounded). Ground-truth (GT) masks use the same region fitting and cover $97\%$ of thin-shadow tanks and nearly all others. Last column: keypoint minus segmentation error with a $95\%$ paired bootstrap interval over tanks.}
\label{tab:width}
\resizebox{\columnwidth}{!}{%
\begin{tabular}{@{}lccccc@{}}
\toprule
\textbf{Width} & \textbf{Tanks} & \textbf{Keypoints} & \textbf{Segm.} & \textbf{GT masks} & \textbf{KP $-$ Segm.} \\
\midrule
$< 6\text{ px}$ & 276 & \textbf{0.0181} & 0.0312 & 0.0262 & $-.013$ {\scriptsize[$-.019$, $-.008$]} \\
$6$--$12\text{ px}$ & 215 & \textbf{0.0138} & 0.0211 & 0.0149 & $-.007$ {\scriptsize[$-.011$, $-.004$]} \\
$\ge 12\text{ px}$ & 743 & 0.0097 & 0.0104 & 0.0062 & $-.001$ {\scriptsize[$-.003$, $+.001$]} \\
\midrule
All & 1,234 & \textbf{0.0123} & 0.0169 & 0.0121 & $-.005$ {\scriptsize[$-.007$, $-.003$]} \\
\bottomrule
\end{tabular}%
}
\end{table}

\begin{figure}[t]
\centering
\includegraphics[width=\columnwidth]{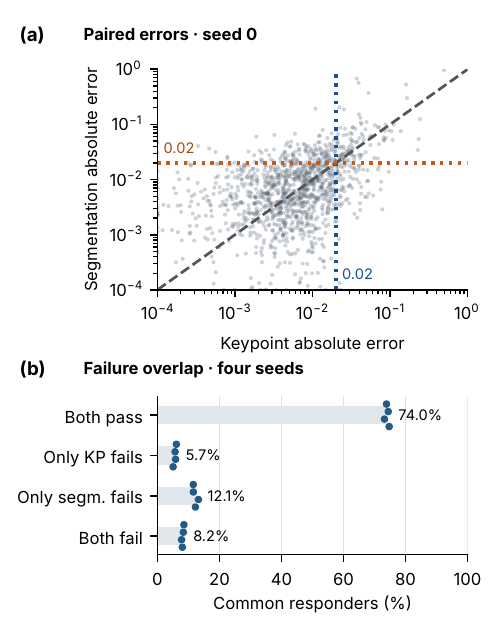}
\caption{Paired affine-estimator errors and failure overlap on test tanks scored by both the keypoint (KP) and matched ResNet-18 U-Net segmentation pipelines. (a) For seed 0 (1,231 tanks), the diagonal marks equal error and dotted lines mark the $0.02$ failure threshold. Errors below $10^{-4}$ are shown at that limit. (b) Across four seeds, bars show mean outcome shares and dots show individual seed shares (1,231--1,235 tanks per seed). Abstentions are excluded; failure means absolute roof-state error $>0.02$.}
\label{fig:kp_vs_seg}
\end{figure}

The gap depends on shadow width (Table~\ref{tab:width}). Below $6\text{ px}$, keypoints have $42\%$ lower error and the paired interval excludes zero. It also excludes zero from $6$ to $12\text{ px}$; at $12\text{ px}$ and above, the observed difference is small.

\subsection{RQ2: Cost of the nadir assumption}
\label{sec:rq2}

\begin{table}[t]
\centering
\caption{Keypoint-pipeline test roof-state MAE by camera elevation ($90^\circ$ is nadir), mean over four seeds, on the tanks it covers ($98$--$100\%$ per band for the nadir estimator). The nadir and affine estimators convert the same predicted keypoints.}
\label{tab:tilt}
\resizebox{\columnwidth}{!}{%
\begin{tabular}{@{}lcccc@{}}
\toprule
\textbf{Camera elevation} & $60^\circ$--$65^\circ$ & $65^\circ$--$75^\circ$ & $75^\circ$--$85^\circ$ & $85^\circ$--$90^\circ$ \\
\midrule
Nadir estimator & 0.147 & 0.122 & 0.073 & 0.022 \\
Affine estimator & \textbf{0.012} & \textbf{0.015} & \textbf{0.012} & \textbf{0.010} \\
MAE ratio (nadir / affine) & $12\times$ & $8\times$ & $6\times$ & $2\times$ \\
Tanks & 201 & 436 & 401 & 211 \\
\bottomrule
\end{tabular}%
}
\end{table}

Converting the same predicted keypoints with the nadir formula raises the error at every camera elevation, from twice that of the affine estimator near nadir to twelve times at $60^\circ$--$65^\circ$, while the affine estimator stays between $0.010$ and $0.015$ (Table~\ref{tab:tilt}). For segmentation, the nadir-to-affine error ratio is $7.7\times$ at $60^\circ$--$65^\circ$; this understates the cost because the nadir estimator abstains on $35\%$ of these tanks (Table~\suppref{tab:supp_tilt}{S8}).

Sun elevation is the other operating limit (Table~\ref{tab:sun}). As the sun rises, the interior shadow shortens with $\cot\theta$, so a fixed localization error becomes a larger depth error: keypoint error stays at $0.004$--$0.008$ below $55^\circ$ and reaches $0.038$ above $70^\circ$, where $47\%$ of estimates fail.

\begin{table}[t]
\centering
\caption{Test roof-state MAE using the affine estimator by solar elevation on the common subset, mean over four seeds (tanks per seed). Keypoints are more accurate in every band and seed.}
\label{tab:sun}
\resizebox{\columnwidth}{!}{%
\begin{tabular}{@{}lccccc@{}}
\toprule
\textbf{Sun elevation} & $10^\circ$--$25^\circ$ & $25^\circ$--$40^\circ$ & $40^\circ$--$55^\circ$ & $55^\circ$--$70^\circ$ & $70^\circ$--$90^\circ$ \\
\midrule
Keypoints & \textbf{0.0040} & \textbf{0.0051} & \textbf{0.0081} & \textbf{0.0158} & \textbf{0.0383} \\
Segmentation & 0.0045 & 0.0070 & 0.0108 & 0.0206 & 0.0590 \\
Tanks & 223 & 244 & 265 & 372 & 130 \\
\bottomrule
\end{tabular}%
}
\end{table}

\subsection{RQ3: Failure conditions and error patterns}
\label{sec:rq3}

\paragraph{Keypoint failures.}
Pooling the 4,997 validation and test tanks of keypoint seed 0 gives 709 failures and 40 abstentions where the shadow is hidden. Clustering the failures ($k$-means, $k=4$ chosen by silhouette) yields four groups with mean sun elevations of $65^\circ$--$69^\circ$. (A) For 166 small tanks seen in oblique views, the near wall hides most of the interior shadow, and predicted shadow points fall outside the visible shadow. (B) For 49 nearly full tanks with a true shadow about $0.5\text{ px}$ wide, predictions include spurious keypoints. (C) For 176 large tanks (median diameter $230\text{ px}$), rim keypoints are off by $2.7\text{ px}$ on average. (D) In the largest group (318 tanks), small shared rim and centre offsets put estimates just beyond the threshold, at the lowest mean error ($0.037$). Groups A and B contain the largest errors (Fig.~\suppref{fig:supp_failure_distribution}{S3}).

\paragraph{Segmentation errors.}
Ground-truth masks provide a reference for the segmentation readout in Table~\ref{tab:width}. The comparison with learned keypoints combines differences in the inputs and readouts; it does not isolate rasterization. On thin shadows, ground-truth masks have $0.008$ higher error than learned keypoints, and predicted masks have a further $0.005$ higher error under region fitting. On wide shadows, ground-truth masks have $0.0035$ lower error than keypoints, while predicted masks have $0.004$ higher error than ground-truth masks. Better mask prediction could close the wide-shadow gap. On thin shadows, a sub-pixel contour readout may improve the segmentation pipeline, but remains to be evaluated.

\paragraph{Failure overlap.}
On test tanks scored by both pipelines, both pass in $74.0\%$ of cases on average across seeds, and both fail in $8.2\%$. Only keypoints fail in $5.7\%$, while only segmentation fails in $12.1\%$ (Fig.~\ref{fig:kp_vs_seg}). These shares exclude abstentions and are computed separately for each seed.

\subsection{Results on real imagery}
\label{sec:detection}

Every fill model is trained on SynTank-25k alone, because real imagery has no fill labels. We measure transfer to NAIP imagery in two steps: detection, which real box labels can score, and keypoint fill estimation, which can only be checked for internal consistency.

\paragraph{Detection.}
The public NAIP storage-tank dataset~\cite{robinson2024remotely} has 27,639 RGB chips of $512\times512\text{ px}$ at $0.6\text{ m/px}$, with bounding boxes in seven categories and acquisition dates, but no fill or height labels. We remove 343 duplicates and split the chips 80/15/5 (21,837 / 4,094 / 1,365), keeping chips that share a USGS quarter-quadrangle, overlap, or are near-duplicates in the same split. We train YOLO26s-OBB detectors on real-only, synthetic-only or mixed data under one protocol (one run per condition, with settings in Appendix~\suppref{supp:detection}{I}) and score all three on the same $1,365$ real test chips and $2,895$ synthetic test tiles. Synthetic training contains only floating-roof and closed-roof tanks, so Table~\ref{tab:det_transfer} averages AP over these two classes and also reports floating-roof AP, the class the fill pipeline uses.

\begin{table}[t]
\centering
\caption{Detection AP averaged over IoU thresholds from 0.50 to 0.95. All training conditions use the same test set within each domain. Shared-class mAP averages AP across floating-roof and closed-roof tanks; floating-roof AP reports the class used for fill estimation.}
\label{tab:det_transfer}
\resizebox{\columnwidth}{!}{%
\begin{tabular}{@{}lcccc@{}}
\toprule
 & \multicolumn{2}{c}{\textbf{Shared-class mAP}} & \multicolumn{2}{c}{\textbf{Floating-roof AP}} \\
\cmidrule(lr){2-3}\cmidrule(lr){4-5}
\textbf{Training data} & \textbf{Real} & \textbf{Synthetic} & \textbf{Real} & \textbf{Synthetic} \\
\midrule
Real only & $0.8767$ & $0.7835$ & $0.8924$ & $0.8415$ \\
Synthetic only & $0.6082$ & $0.9708$ & $0.6164$ & $0.9756$ \\
Mixed & $0.8767$ & $0.9686$ & $0.8970$ & $0.9735$ \\
\bottomrule
\end{tabular}%
}
\end{table}

The synthetic-only detector finds $93\%$ of the 527 real floating-roof test tanks with the correct class (box IoU $\ge 0.5$, confidence $\ge 0.25$), compared with $89\%$ for the real-trained detector, and it misses fewer small tanks. The synthetic-only detector produces 6,557 false positives on the real test chips, six times the real-trained detector's count. Most occur on vehicles, rail cars, rooftop units and treatment basins, which are absent from the synthetic images. Mixed training matches real-only training on real imagery and approaches synthetic-only performance on synthetic tiles.

\begin{figure*}[t]
\centering
\includegraphics[width=0.65\textwidth]{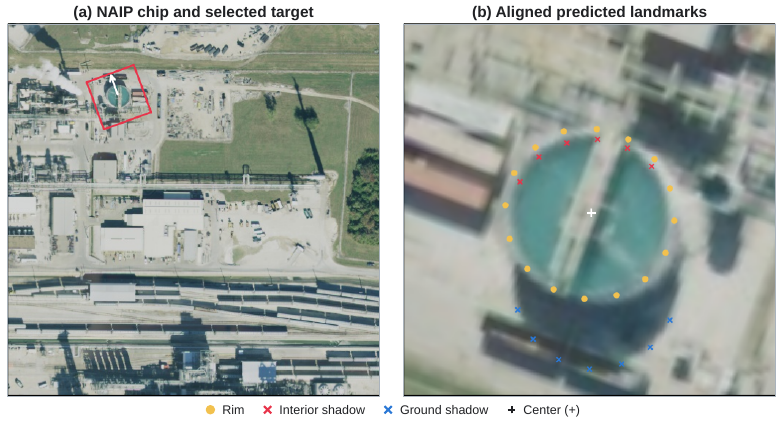}
\caption{Keypoint predictions on a real NAIP chip ($512\times512$ px at $0.6$ m/pixel), using an annotated box for a tank passing the consistency checks. (a) Chip with the selected crop in red and inferred shadow direction in white. (b) Crop aligned by a rotation search over shadow directions allowed by acquisition date and location, with keypoints predicted by the synthetic-trained model.}
\label{fig:naip_prediction_example}
\end{figure*}

\paragraph{Fill-estimate consistency.}
\label{sec:real}
We apply the synthetic-trained keypoint model (seed 0) to $6,558$ complete NAIP floating-roof annotations, excluding $3,640$ border-cut tanks and $97$ with no-data pixels. Crops use annotated boxes, so this analysis isolates the keypoint and physics stages from detection; $336$ of the tanks lie in held-out detector test chips. Figure~\ref{fig:naip_prediction_example} shows a prediction on a real chip.

NAIP provides acquisition dates and locations, but capture times and camera angles are unavailable. For each crop, we search rotations within the solar-direction corridor allowed by date and location, infer solar elevation from the selected direction, and assume a nadir camera. Under this assumption, the parallax term $b$ is zero, so the NAIP analysis evaluates the nadir case. The affine estimator normalizes depth with a diameter-indexed wall-height prior ($12$, $14$, $16$ and $22\text{ m}$ at diameters of $20$, $40$, $60$ and $80\text{ m}$, interpolated linearly and held constant outside that range).

\begin{figure}[t]
\centering
\includegraphics[width=\linewidth]{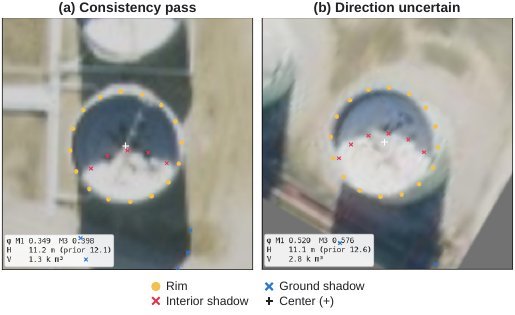}
\caption{Real NAIP crops with predicted keypoints: (a) median-disagreement case among those passing the checks; (b) median-severity case flagged for uncertain shadow direction. Crops are aligned with the inferred shadow direction pointing down; white $+$ marks the predicted center. Insets report roof state from M1 (displacement ratio) and M3 (affine), ground-shadow height H, and M1-based volume V estimates, with the height prior in parentheses. NAIP fill ground truth is unavailable.}
\label{fig:naip_screen_examples}
\end{figure}

The affine estimator returns an estimate for $98\%$ of these tanks and the displacement-ratio estimator for $95\%$. A tank passes the consistency checks when both shadow boundaries are found, the inferred shadow direction agrees with the solar geometry, the implied wall height is plausible, and the displacement-ratio and affine estimates agree (thresholds in Appendix~\suppref{supp:naip}{H}). In total, $3,261/6,558$ tanks ($49.7\%$) pass, including $150/336$ ($44.6\%$) in held-out test chips. The median absolute difference between the two estimates is $0.049$ among passing tanks and $0.104$ among all tanks with both estimates. These rates summarize internal consistency: the two estimators share predicted shadow measurements, and the checks select partly on their agreement. Figure~\ref{fig:naip_screen_examples} contrasts a passing case with one flagged for uncertain shadow direction; Fig.~\suppref{fig:supp_naip_gallery}{S5} shows the other outcome categories.


\section{Discussion}
\label{sec:discussion}

\paragraph{Takeaways.}
Fill-error reports should include shadow width, because the keypoint--segmentation gap grows as shadows narrow (Table~\ref{tab:width}). On a $15\text{ m}$ wall, the overall gap corresponds to about $0.07\text{ m}$ of roof depth and the thin-shadow gap to about $0.20\text{ m}$. Camera geometry matters more: at $60^\circ$--$65^\circ$ camera elevation, the nadir conversion of the same keypoints errs by about $2\text{ m}$ more than the affine estimator. High sun shortens interior shadows, weakening the depth signal.

\paragraph{Limitations.}
\textit{Real-image evidence.} We measure fill, roof-height, and volume errors on synthetic tanks. On NAIP, inferred solar geometry, the nadir-camera assumption, and a height prior limit the results to consistency checks. NAIP is flown during the agricultural growing season~\cite{robinson2024remotely}, when high sun can shorten the shadow. Orthorectification may leave relief displacement of elevated tank walls, so the nadir assumption may not hold exactly. Real-fill accuracy requires gauge records matched to acquisition times. Evaluating the oblique correction on real imagery requires off-nadir views with reliable acquisition geometry.

\textit{Simulation and acquisition.} The synthetic scenes use an orthographic camera and exact solar metadata and wall heights. The imagery omits higher-order sensor distortions, and reported errors exclude metadata error. At half depth on a $15\text{ m}$ wall, a $1\text{ m}$ wall-height error shifts estimated fill by about $0.03$. A $1^\circ$ error in a $45^\circ$ solar elevation shifts estimated fill by about $0.018$. Both effects exceed the overall pipeline gap. Other real-world error sources include roof and seal construction, terrain height, and ambiguous shadow edges.

\textit{Generalization and model comparison.} The scene-group split measures performance within the sampled design. Testing transfer to unseen sites or tank-design ranges requires holding them out. Synthetic tanks also differ from NAIP tanks in diameter ($12$--$90\text{ m}$ vs. $3$--$69\text{ m}$) and median image-space diameter ($99$ vs. $56\text{ px}$). Width strata are measured in render pixels; in the zoomed half, effective sensor resolution is coarser. The two pipelines share backbone, input size and crop, but use different readouts: displacement fitting for keypoints and region fitting for masks. An untested sub-pixel contour readout for soft masks could narrow the thin-shadow gap.

\section{Conclusion}
\label{sec:conclusion}

SynTank-25k provides exact geometric labels for evaluating floating-roof fill estimation from overhead imagery. A detector trained on SynTank-25k recovers 93\% of floating-roof test tanks in NAIP. Modelling camera parallax reduces the nadir conversion's error up to twelvefold on oblique views. When backbone, input size and crop are matched, the keypoint pipeline has lower roof-state error than the segmentation pipeline, mainly on thin shadows. Ground-truth masks converted by region fitting also have higher error than learned keypoints in this band. For wide shadows, differences are small; better mask predictions could close the gap.

{
    \small
    \bibliographystyle{ieeenat_fullname}
    \bibliography{main}
}

%% file: supplement.tex
\maketitle

\appendix
\setcounter{figure}{0}
\setcounter{table}{0}
\renewcommand{\thefigure}{S\arabic{figure}}
\renewcommand{\thetable}{S\arabic{table}}
\renewcommand{\theHfigure}{supp.\arabic{figure}}
\renewcommand{\theHtable}{supp.\arabic{table}}

\section*{Contents}
{\renewcommand{\arraystretch}{1.15}
\noindent\begin{tabular}{@{}p{0.18\columnwidth}p{0.66\columnwidth}r@{}}
\toprule
Section & Contents & Page \\
\midrule
\rowcolor{wacvblue!6}\hyperref[supp:geometry]{\S\ref*{supp:geometry}} & Geometry and solver & \pageref{supp:geometry} \\
\hyperref[supp:dataset]{\S\ref*{supp:dataset}} & Dataset design and labels & \pageref{supp:dataset} \\
\rowcolor{wacvblue!6}\hyperref[supp:validation]{\S\ref*{supp:validation}} & Validation and splits & \pageref{supp:validation} \\
\hyperref[supp:training]{\S\ref*{supp:training}} & Models and training & \pageref{supp:training} \\
\rowcolor{wacvblue!6}\hyperref[supp:metrics]{\S\ref*{supp:metrics}} & Metrics, seeds, and coverage & \pageref{supp:metrics} \\
\hyperref[supp:true]{\S\ref*{supp:true}} & Results with ground-truth inputs & \pageref{supp:true} \\
\rowcolor{wacvblue!6}\hyperref[supp:envelope]{\S\ref*{supp:envelope}} & Operating envelope and failures & \pageref{supp:envelope} \\
\hyperref[supp:naip]{\S\ref*{supp:naip}} & Real NAIP protocol and diagnostics & \pageref{supp:naip} \\
\rowcolor{wacvblue!6}\hyperref[supp:detection]{\S\ref*{supp:detection}} & Detector transfer & \pageref{supp:detection} \\
\bottomrule
\end{tabular}}

\section*{Notation and evaluation conventions}
Unless stated otherwise, \(\phi_R\) denotes the ground-referenced roof state. We report test MAE over valid predictions and give the number of scored tanks beside each value. By default, pipeline comparisons use test tanks that both pipelines score under the given estimator; tables that use each pipeline's own coverage state this explicitly. ``Segmentation'' denotes the ResNet-18 U-Net matched to the keypoint model; we name the smaller MobileNetV3 U-Net explicitly. ``Seed'' refers to model initialization (0--3); seed 46 is the SynTank-25k generation seed.

\section{Geometry and implemented solver}
\label{supp:geometry}

Let the local orthographic projection over a tank be \(y=Ax+bz+c\), where \(x\) is horizontal ground position, \(z\) is elevation, and \(y\) is image position in pixels. The vector \(b\) is the image displacement per metre of elevation caused by off-nadir viewing. If \(\hat{s}\) is the horizontal anti-solar direction and \(\theta\) the solar elevation, the roof shadow is displaced horizontally by \(\Delta_R\cot\theta\hat{s}\). Its projected displacement from the inner rim to the shadow boundary is
\begin{equation}
  \delta_1=\Delta_R a_s,\qquad a_s=A(\cot\theta\,\hat{s})-b.
  \label{eq:supp_shadow}
\end{equation}
The ground shadow has displacement \(\delta_3=H_Ga_s\). With \(\Delta_R=z_C-z_R\) and \(H_G=z_C-z_G\), the reported state is
\begin{equation}
  \phi_R=\frac{z_R-z_G}{H_G}=1-\frac{\Delta_R}{H_G}.
  \label{eq:supp_roof_state}
\end{equation}
In SynTank-25k the tank floor, which is also the lowest deck position, lies at \(z_F=0.2\) m above the ground. Consequently, \(\phi_R\) differs slightly from the liquid fill fraction \((z_R-z_F)/(z_C-z_F)\), which it equals only when floor and ground elevations coincide; volume uses the liquid height \(z_R-z_F\), with \(V=\pi R^2\max(z_R-z_F,0)\).

The estimators use different observations and acquisition metadata: the shadow-displacement ratio (M1), nadir estimator (M2), and affine estimator (M3). M1 estimates \(\phi_R=1-\|\delta_1\|/\|\delta_3\|\). The factor \(a_s\) cancels, but M1 requires both inner and exterior ground-shadow measurements. M2 converts the inner displacement to depth using GSD and solar elevation, assuming \(b=0\) (nadir camera). M3 searches over roof elevation, fitting the observed inner-shadow displacement vectors to \((z_C-z_R)a_s\) with a 1-pixel Huber loss to reduce sensitivity to keypoint outliers. It searches roof elevations on a 0.005 m grid and applies parabolic refinement around the minimum. We report the median residual in pixels.

The keypoint model predicts a centre and 16 sun-relative keypoints on each of four curves: rim crest, inner shadow boundary, exterior ground-shadow boundary, and roof edge. Each index uses the same sun-relative world azimuth to construct the corresponding points; shadow points are displaced from their casters. For each index, the inner-rim caster is interpolated from its rim keypoint using the known shell thickness. A keypoint is usable when its predicted visibility probability exceeds 0.5. M3 abstains when no inner-shadow keypoint is usable; M1 also requires an exterior-shadow keypoint. The segmentation pipeline uses the same physical model with a region-based mask readout. At each candidate roof elevation, the affine model predicts the interior-shadow region inside the rim lip. The readout compares that region with the predicted soft inner-shadow probability using a \(4\times4\) sub-pixel coverage grid per crop pixel. It searches over 0.10 m and refines at 0.01 m. The 16 samples approximate the coverage integral. The fit abstains when the summed inner-shadow probability inside the lip is below 10 crop pixels. For M1 and M2, we estimate inner displacement by shifting the rim lip to align with the predicted inner shadow. We estimate exterior displacement the same way from the ground shadow. A displacement is undefined when lip evidence is below 10 crop pixels or no shift reduces mismatch by at least 10\% relative to zero shift. M1 also requires a measurable ground shadow whose tip lies inside the crop mosaic. These gates explain the lower M2 coverage of segmentation; most abstentions occur on oblique views of thin shadows. We clip reported fill estimates to \([0,1]\).

\section{SynTank-25k design and labels}
\label{supp:dataset}

SynTank-25k is the accepted subset of a 25,000-image Blender plan generated with seed 46. It contains 24,987 images arranged in 18,608 scene groups. The plan varies tank radius (6--45 m), shell height (9--22 m), continuous roof state, five sites, six layouts, three atmospheric conditions, sun bands, and camera elevation (60--90 degrees). The site counts after exclusions are shown in Table~\ref{tab:supp_sites}. Scene-level labels provide physical elevations, tank dimensions, sun and camera geometry, and projected coordinates. The keypoint target contains 65 points with visibility labels. The same scenes also provide interior and exterior shadow masks and detection boxes. The masks, keypoints, and boxes are generated from the same synthetic scenes.

The five sites are Ras Tanura, Zhoushan, Cushing, Primorsk, and Rotterdam. Acquisition windows are 07:30--16:30 local time for most sites, 09:30--11:30 at Primorsk, and 09:00--13:00 at Rotterdam. Planned sun-elevation bands span $10^\circ$--$90^\circ$. The rim shadow covers no more than $85\%$ of the roof diameter, and camera azimuth is resampled until at least $10\%$ of the roof is visible and sunlit. Seal gaps range from $0.05$ to $0.35\text{ m}$, and deck tilt is at most $0.5^\circ$.

Half of the synthetic images use NAIP-scale sampling directly; the remaining half use log-uniform $1$--$4\times$ zoom, with a coarsest render sampling distance of $0.8\text{ m/px}$. Each $800\times600$ synthetic image passes through optical blur, resampling, simulated Brovey-style pan-sharpening, shot and read noise, radiometric variation, sharpening, and JPEG compression before returning to the original size. The effective ground sample distance is $0.5$--$0.7\text{ m}$. Shadow-width strata in the paper are measured in render pixels, so a pixel in a zoomed synthetic image is not always one effective sensor pixel. NAIP's public tank inventory supplies boxes, diameters, and acquisition dates~\cite{robinson2024remotely}. We use real imagery for detection and a consistency analysis; fill accuracy is measured on synthetic images with exact truth.

\begin{figure*}[t]
\centering
\includegraphics[width=\textwidth]{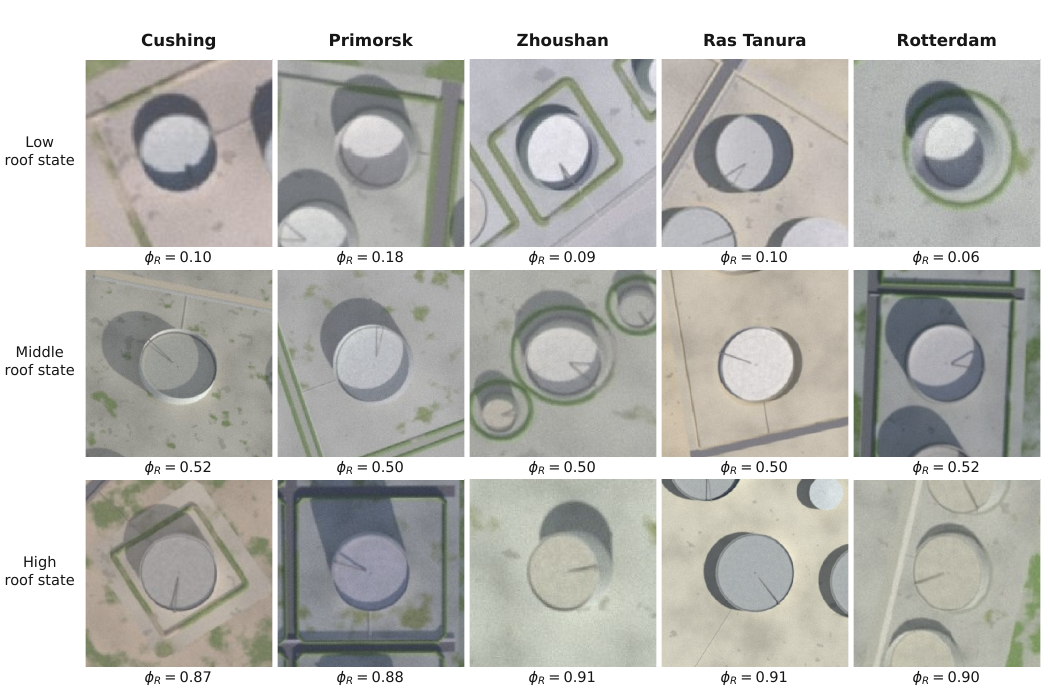}
\caption{SynTank-25k examples from the held-out test set, restricted to tanks of radius at least 15 m. Columns identify generator site presets. Rows show low (\(0\leq\phi_R<0.2\)), middle (\(0.4\leq\phi_R<0.6\)), and high (\(0.8\leq\phi_R<1\)) roof states; values beneath crops give ground-referenced \(\phi_R\), computed from the labeled elevations using Eq.~\ref{eq:supp_roof_state}. Each square crop is centered on the annotated rim center and spans 4.8 tank radii at render sampling. Crops are displayed at a common size, without a shared ground scale. Other scene parameters vary across examples; the figure illustrates tank appearance across roof states.}
\label{fig:supp_syntank_examples}
\end{figure*}

Figure~\ref{fig:supp_syntank_examples} illustrates the rendered tank appearance across generator site presets and roof-state bands. Within each site and band, we select the test image closest to \(\phi_R=0.1\), \(0.5\), or \(0.9\), respectively, requiring a tank radius of at least 15 m, a crop width of at least 100 render pixels, and no crop clipping at the image boundary. The gallery emphasizes readable examples, so its site and size frequencies reflect that selection.

\begin{table}[t]
\centering
\caption{Accepted SynTank-25k images by generator site after validation exclusions.}
\label{tab:supp_sites}
\small
\begin{tabular}{@{}lr@{}}
\toprule
Site & Images \\
\midrule
China & 5,997 \\
Netherlands & 1,000 \\
Russia & 5,998 \\
Saudi Arabia & 5,995 \\
United States & 5,997 \\
\midrule
Total & 24,987 \\
\bottomrule
\end{tabular}
\end{table}

\section{Validation and split integrity}
\label{supp:validation}

The image validator checks image and mask dimensions, consistency of label geometry, roof brightness, and agreement of analytic keypoint curves. Its shadow-separation test offsets candidate inner-shadow points by 1.5 pixels and checks whether the samples cross from shadow to lit roof. At least eight points must be testable, and at least 95\% of testable points must separate correctly. The test was untestable for 2,076 synthetic images; the validator excludes these cases from its failure count. The final report records 24,987 passes and 13 failures, with the 13 excluded filenames listed in the generation directory.

One shared split assigns complete scene groups to train, validation, and test: 19,990 / 3,748 / 1,249 synthetic images in 13,927 / 3,455 / 1,226 scene groups, respectively. The image audit found no byte-identical or near-duplicate cross-split pairs. It flags candidate pairs using \(16\times16\) grayscale thumbnails with correlation above 0.99, then confirms them when the \(128\times128\) versions differ by less than 2 grey levels on average. The maximum cross-split image correlation was 0.978.

Metadata flags identify two radius--height ranges, low-sun scenes, and Rotterdam as potential held-out conditions. Tagged scenes occur in every split, so the results describe performance on the shared in-distribution split.

\section{Models and training protocol}
\label{supp:training}

The keypoint and segmentation models use the shared synthetic split and were trained for 35 epochs with four seeds (0--3). The keypoint crop spans four tank radii and is resampled to 256\(\times\)256 pixels; its ResNet-18 SimpleBaseline predicts 65 heatmaps at 128\(\times\)128 resolution and a visibility head. Its losses combine group-weighted heatmap MSE, visibility BCE, coordinate smooth-L1, and displacement smooth-L1. Coordinate and displacement losses are gated until the heatmaps form and ramp over three epochs. Augmentation uses scale changes up to 10\%, rotations up to 5 degrees, shifts up to 5\%, and horizontal flips with probability 0.5.

\begin{figure*}[t]
\centering
\includegraphics[width=0.98\textwidth]{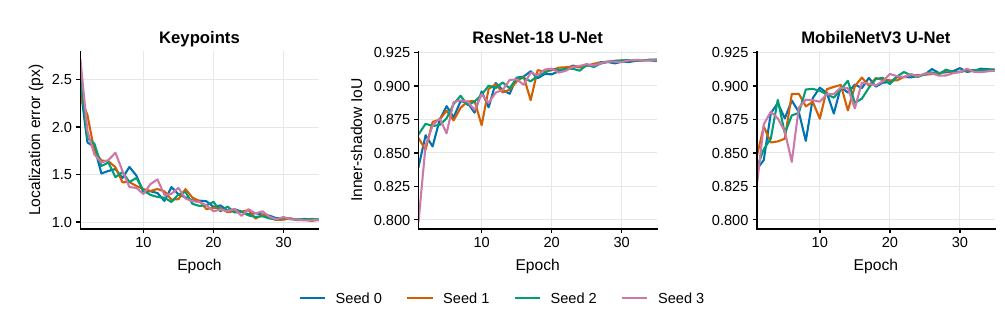}
\caption{Validation learning curves for four seeds. Left: keypoint localization error. Centre: inner-shadow IoU for the matched ResNet-18 U-Net. Right: inner-shadow IoU for the smaller MobileNetV3 U-Net. Checkpoints minimize keypoint localization error or maximize inner-shadow IoU, respectively. The segmentation models use different crop scales.}
\label{fig:supp_learning}
\end{figure*}

The segmentation baseline is an ImageNet-pretrained ResNet-18 U-Net with background, ground-shadow, and inner-shadow classes. Its crop is 256\(\times\)256 pixels and spans four tank radii, so backbone, input size and crop match the keypoint model. A smaller ImageNet-pretrained MobileNetV3-Small U-Net~\cite{mobilenetv3} on 128\(\times\)128 crops spanning three radii is trained with the same protocol. Both segmentation networks use class-weighted cross-entropy plus soft Dice loss on the two shadow classes. Table~\ref{tab:supp_train} summarizes the training configurations and checkpoint-selection criteria, and Fig.~\ref{fig:supp_learning} shows the corresponding validation curves for the keypoint model, matched baseline and smaller model. All models use AdamW with learning rate \(10^{-3}\) and weight decay \(10^{-4}\); each uses its own validation metric for checkpoint selection. The four seed runs share the same test tanks.

\paragraph{Keypoint decoding.} Each of the 65 heatmaps has \(128\times128\) cells. A keypoint is decoded at the argmax cell and refined by the distribution-aware second-order Taylor step of Zhang et al.~\cite{9157744}, computed on the logarithm of the heatmap after an \(11\times11\) Gaussian smoothing. The offset is clipped to one cell, and refinement is skipped within two cells of the heatmap border. Cell coordinates map to input pixels as \((x+0.5)\,s-0.5\) with stride \(s=2\). A keypoint is usable when its predicted visibility probability exceeds 0.5.

\begin{table*}[t]
\centering
\caption{Model and training details. The ResNet-18 U-Net matches the keypoint model in backbone, input size and crop extent; the two differ in output representation and in how fill is read out.}
\label{tab:supp_train}
\small
\resizebox{\textwidth}{!}{%
\begin{tabular}{@{}p{0.13\textwidth}p{0.27\textwidth}p{0.27\textwidth}p{0.27\textwidth}@{}}
\toprule
 & Keypoint pipeline & Segmentation (ResNet-18 U-Net) & MobileNetV3 U-Net \\
\midrule
Backbone / head & ImageNet ResNet-18 SimpleBaseline; 65 heatmaps and visibility head & ImageNet ResNet-18 U-Net; 3 classes & ImageNet MobileNetV3-Small U-Net; 3 classes \\
Parameters & 16.5M & 13.6M & 0.96M \\
Crop / output & Four radii; 256\(\times\)256 crop; 128\(\times\)128 heatmaps & Four radii; 256\(\times\)256 mask & Three radii; 128\(\times\)128 mask \\
Loss & Weighted heatmap MSE, visibility BCE, coordinate smooth-L1, displacement smooth-L1 & Class-weighted cross-entropy and soft Dice on shadow classes & Same as ResNet-18 U-Net \\
Optimization & AdamW, \(10^{-3}\), weight decay \(10^{-4}\); 35 epochs; batch 32 & AdamW, \(10^{-3}\), weight decay \(10^{-4}\); 35 epochs; batch 16 & AdamW, \(10^{-3}\), weight decay \(10^{-4}\); 35 epochs; batch 16 \\
Checkpoint & Minimum validation keypoint localization error & Maximum validation inner-shadow IoU & Maximum validation inner-shadow IoU \\
Seeds & 0, 1, 2, 3 & 0, 1, 2, 3 & 0, 1, 2, 3 \\
\bottomrule
\end{tabular}%
}
\end{table*}

\section{Test metrics, seeds, and coverage}
\label{supp:metrics}

Table~\ref{tab:supp_seed_results} gives, for each seed, the test roof-state error of the keypoint and segmentation pipelines on the tanks both score under each estimator. The keypoint pipeline has the lower error for every estimator in every seed. The main paper reports the four-seed mean and sample standard deviation of these values. Its width intervals come from a paired bootstrap over tanks: each test tank carries its keypoint-minus-segmentation differences from all seeds, and tanks are resampled 2,000 times. Seed variation and the bootstrap answer different questions: the first concerns model initialization, the second the sampling of test tanks.

\begin{table*}[t]
\centering
\caption{Per-seed test roof-state MAE on the tanks both the keypoint and segmentation pipelines score under each estimator; \(n\) is that number of tanks out of 1,249.}
\label{tab:supp_seed_results}
\small
\resizebox{\textwidth}{!}{%
\begin{tabular}{@{}lrrrrrrrrr@{}}
\toprule
 & \multicolumn{3}{c}{Affine (M3)} & \multicolumn{3}{c}{Displacement ratio (M1)} & \multicolumn{3}{c}{Nadir (M2)} \\
\cmidrule(lr){2-4}\cmidrule(lr){5-7}\cmidrule(lr){8-10}
Seed & \(n\) & Keypoints & Segm. & \(n\) & Keypoints & Segm. & \(n\) & Keypoints & Segm. \\
\midrule
0 & 1,231 & 0.01214 & 0.01623 & 730 & 0.01348 & 0.02055 & 1,023 & 0.07470 & 0.07836 \\
1 & 1,235 & 0.01217 & 0.01588 & 725 & 0.01291 & 0.02107 & 1,025 & 0.07479 & 0.07861 \\
2 & 1,235 & 0.01264 & 0.01855 & 719 & 0.01273 & 0.02077 & 1,019 & 0.07409 & 0.07800 \\
3 & 1,234 & 0.01216 & 0.01711 & 725 & 0.01400 & 0.01979 & 1,017 & 0.07446 & 0.07874 \\
\midrule
Mean \(\pm\) SD & & 0.0123 \(\pm\) 0.0002 & 0.0169 \(\pm\) 0.0012 & & 0.0133 \(\pm\) 0.0006 & 0.0205 \(\pm\) 0.0005 & & 0.0745 \(\pm\) 0.0003 & 0.0784 \(\pm\) 0.0003 \\
\bottomrule
\end{tabular}%
}
\end{table*}

Table~\ref{tab:supp_own} reports each pipeline's error on the tanks it covers. Because the pipelines abstain on different tanks, their own-coverage MAEs reflect different test sets. MobileNetV3 abstains more often on difficult thin-shadow tanks, lowering its own-coverage MAE relative to the ResNet-18 U-Net. On tanks scored by all three pipelines, both segmentation networks have affine-estimator error 0.0158; keypoints have error 0.0116.

\begin{table*}[t]
\centering
\caption{Per-seed test MAE using the affine estimator with each pipeline scored on the tanks it covers (scored / abstained among 1,249 tanks).}
\label{tab:supp_own}
\small
\begin{tabular}{@{}lrrr@{}}
\toprule
Seed & Keypoints & Segmentation (ResNet-18 U-Net) & MobileNetV3 U-Net \\
\midrule
0 & 0.01263 (1,236 / 13) & 0.01689 (1,235 / 14) & 0.01656 (1,225 / 24) \\
1 & 0.01248 (1,239 / 10) & 0.01730 (1,240 / 9) & 0.01702 (1,229 / 20) \\
2 & 0.01351 (1,243 / 6) & 0.01871 (1,237 / 12) & 0.01644 (1,230 / 19) \\
3 & 0.01263 (1,240 / 9) & 0.01773 (1,237 / 12) & 0.01577 (1,225 / 24) \\
\midrule
Mean \(\pm\) SD & 0.01281 \(\pm\) 0.00047 & 0.01766 \(\pm\) 0.00078 & 0.01645 \(\pm\) 0.00052 \\
\bottomrule
\end{tabular}
\end{table*}


For keypoint seed 0, the pipeline scored 1,236 test tanks and abstained on 13. On these tanks, its median absolute error was 0.00646, signed bias was +0.00131, roof-height MAE was 0.192 m, and volume error was 1.28\% of tank capacity. The within-0.02 and within-0.05 rates were 85.1\% and 96.4\% of covered tanks, respectively. Averaged over seeds, the segmentation network's test IoU is 0.915 for the interior shadow and 0.935 for the ground shadow (MobileNetV3 U-Net: 0.907 and 0.921). Boundary F1 uses a crop-pixel tolerance, and the two networks use different crop scales, so their scores are not directly comparable. IoU and Boundary F1 quantify mask quality, while roof-state error measures estimation accuracy.

\section{Results with ground-truth inputs}
\label{supp:true}

The references replace network outputs with synthetic ground truth, then apply the same conversion and validity rules. They measure the error of the readouts tested here with ground-truth inputs. The mask reference uses region fitting; a sub-pixel contour readout remains to be evaluated.

On the common subset, ground-truth keypoints give an affine-estimator error of 0.0016 and ground-truth masks at the segmentation baseline's crop give 0.0121, each averaged over the common-subset tanks it covers (about 1,223 and 1,224 tanks per seed). Scored on the tanks they cover, ground-truth masks give 0.01320 on 1,231 of 1,249 test tanks at 256 pixels and four radii, and 0.01355 on 1,224 tanks at 128 pixels and three radii. Table~\ref{tab:supp_oracle} lists these values and Table~\ref{tab:supp_width} breaks the comparison down by shadow width, adding the MobileNetV3 U-Net on the common-subset tanks it covers.

\begin{table}[t]
\centering
\caption{Ground-truth input references for the affine estimator on the test split.}
\label{tab:supp_oracle}
\small
\resizebox{\columnwidth}{!}{%
\begin{tabular}{@{}llrr@{}}
\toprule
Input & Tanks & Scored & MAE \\
\midrule
Ground-truth keypoints & Common subset & \(\approx\)1,223 & 0.0016 \\
Ground-truth masks, 256 px & Common subset & \(\approx\)1,224 & 0.0121 \\
Ground-truth masks, 256 px & Own coverage & 1,231 & 0.01320 \\
Ground-truth masks, 128 px & Own coverage & 1,224 & 0.01355 \\
\bottomrule
\end{tabular}%
}
\end{table}

\begin{table*}[t]
\centering
\caption{Test roof-state MAE using the affine estimator by true interior-shadow width (render pixels), mean \(\pm\) SD over four seeds, on the common subset (test tanks covered by both the keypoint and segmentation pipelines). Tank counts are mean per seed, rounded. The MobileNetV3 U-Net and ground-truth masks are averaged over the common-subset tanks they cover (coverage in parentheses).}
\label{tab:supp_width}
\small
\begin{tabular}{@{}lrllll@{}}
\toprule
Width & Tanks & Keypoints & Segmentation & MobileNetV3 U-Net & Ground-truth masks, 256 px \\
\midrule
\(<6\) px & 276 & 0.0181 \(\pm\) 0.0008 & 0.0312 \(\pm\) 0.0020 & 0.0281 \(\pm\) 0.0020 (97\%) & 0.0262 (97\%) \\
6--12 px & 215 & 0.0138 \(\pm\) 0.0010 & 0.0211 \(\pm\) 0.0028 & 0.0183 \(\pm\) 0.0019 (100\%) & 0.0149 (100\%) \\
\(\geq12\) px & 743 & 0.0097 \(\pm\) 0.0005 & 0.0104 \(\pm\) 0.0006 & 0.0107 \(\pm\) 0.0009 (100\%) & 0.0062 (100\%) \\
All & 1,234 & 0.0123 \(\pm\) 0.0002 & 0.0169 \(\pm\) 0.0012 & 0.0158 \(\pm\) 0.0005 (99\%) & 0.0121 (99\%) \\
\bottomrule
\end{tabular}
\end{table*}

\section{Operating envelope and failure accounting}
\label{supp:envelope}

The displacement relation becomes less informative when the shadow is short. Table~\ref{tab:supp_sun} reports affine-estimator error by solar elevation on the common subset, with the MobileNetV3 U-Net on the tanks it covers. The rise at high sun elevation coincides with reduced shadow length and describes this synthetic test distribution. Real-world calibration requires labeled real imagery.

\begin{table}[t]
\centering
\caption{Test MAE by solar elevation using the affine estimator, averaged over four seeds. Keypoints and segmentation use their common subset, with mean tank counts rounded. MobileNetV3 uses its own scored tanks. Its coverage is relative to all test tanks in each band.}
\label{tab:supp_sun}
\small
\begin{tabular}{@{}lrlll@{}}
\toprule
Sun elevation & Tanks & Keypoints & Segm. & MobileNetV3 \\
\midrule
10--25$^\circ$ & 223 & 0.0040 & 0.0045 & 0.0049 (100\%) \\
25--40$^\circ$ & 244 & 0.0051 & 0.0070 & 0.0077 (100\%) \\
40--55$^\circ$ & 265 & 0.0081 & 0.0108 & 0.0110 (100\%) \\
55--70$^\circ$ & 372 & 0.0158 & 0.0206 & 0.0203 (98\%) \\
70--90$^\circ$ & 130 & 0.0383 & 0.0590 & 0.0532 (91\%) \\
\bottomrule
\end{tabular}
\end{table}

Table~\ref{tab:supp_tilt} gives the cost of the nadir formula for each pipeline. Within each pipeline, the nadir (M2) and affine (M3) estimators use the same model predictions. Each error is averaged over the tanks its estimator covers, so the two errors can use different subsets. Segmentation M2 abstains on about a third of the steepest views, so the segmentation ratios understate the cost.

\begin{table*}[t]
\centering
\caption{Test M2 (nadir) and M3 (affine) MAE by camera elevation, mean over four seeds, each pipeline on the tanks it covers. Ratio is M2 / M3; coverage is the M2 share of the band's test tanks (201, 436, 401 and 211 tanks).}
\label{tab:supp_tilt}
\small
\resizebox{\textwidth}{!}{%
\begin{tabular}{@{}lllll@{}}
\toprule
Pipeline & 60--65$^\circ$ & 65--75$^\circ$ & 75--85$^\circ$ & 85--90$^\circ$ \\
\midrule
Keypoints & 0.147 / 0.012, 12.4\(\times\) (98\%) & 0.122 / 0.015, 8.0\(\times\) (99\%) & 0.073 / 0.012, 6.2\(\times\) (100\%) & 0.022 / 0.010, 2.1\(\times\) (100\%) \\
Segmentation & 0.128 / 0.017, 7.7\(\times\) (65\%) & 0.099 / 0.022, 4.6\(\times\) (73\%) & 0.073 / 0.016, 4.5\(\times\) (92\%) & 0.025 / 0.013, 1.9\(\times\) (97\%) \\
MobileNetV3 U-Net & 0.125 / 0.015, 8.4\(\times\) (64\%) & 0.098 / 0.020, 4.9\(\times\) (74\%) & 0.071 / 0.015, 4.7\(\times\) (91\%) & 0.025 / 0.013, 1.8\(\times\) (97\%) \\
\bottomrule
\end{tabular}%
}
\end{table*}

In the seed-0 pooled validation and test analysis, 749 of 4,997 tanks failed or abstained: 709 scored failures (absolute roof-state error above 0.02) and 40 correct abstentions for hidden shadows. We pooled seed-0 validation and test failures to obtain enough cases for k-means clustering, which grouped the 709 failures into four descriptive clusters: A, wall occlusion with spurious inner-shadow keypoints (166); B, spurious keypoints on nearly full tanks (49); C, large tanks with larger rim-keypoint errors (176); and D, the remaining group (318), with a mean absolute error of 0.037. The ground-truth-mask reference in Table~\ref{tab:supp_width} and the main paper analyze segmentation errors.

\begin{figure*}[t]
\centering
\includegraphics[width=\textwidth]{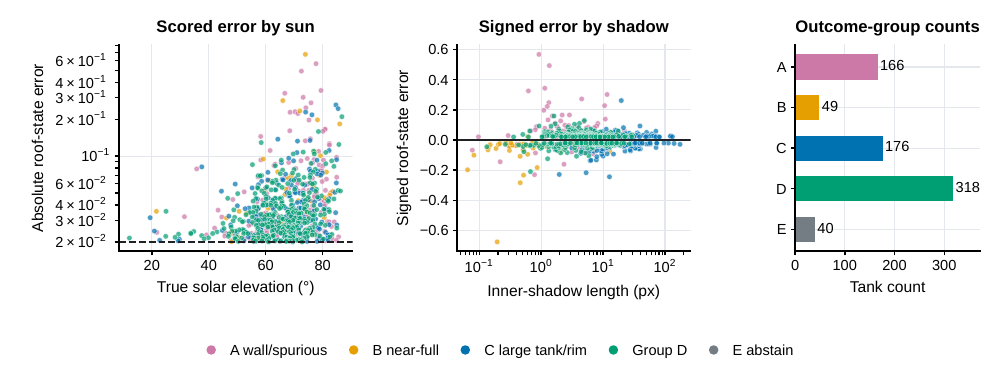}
\caption{Seed-0 failure accounting on pooled validation and test data: 709 scored errors and 40 correct abstentions among 4,997 assigned tanks. A--D are descriptive clusters; E denotes correct abstention. Signed roof-state error is \(\hat{\phi}_R-\phi_R\). Scatter and bar colors share the group key below the panels. The dashed line marks the 0.02 tolerance.}
\label{fig:supp_failure_distribution}
\end{figure*}

\section{Real NAIP application protocol and diagnostics}
\label{supp:naip}

The seed-0 keypoint model trained on SynTank-25k was applied without retraining to 6,558 complete real floating-roof annotations. Annotated tank boxes supply the crops, so this analysis covers the keypoint and physics stages; detector performance is reported separately. The set excludes 3,640 border-cut annotations and 97 annotations whose crops contain no-data pixels; 336 included tanks lie in detector test chips. NAIP supplies location and acquisition date~\cite{robinson2024remotely}. We therefore report application behavior and estimator consistency.

For each crop, the pipeline searches rotations within the solar-direction corridor allowed by date and location. It estimates solar elevation from the selected direction and assumes a nadir orthophoto. M3 estimates roof depth \(\Delta_R\) in metres from shadow geometry. We convert depth to normalized roof state using \(\phi_R=1-\Delta_R/H_G\), where \(H_G\) is a diameter-indexed shell-height prior. The prior is interpolated linearly between the knots \((20,12),(40,14),(60,16),(80,22)\) in metres (diameter, height) and held constant outside that range. This analysis uses nadir geometry; the oblique correction is evaluated on synthetic test images. M1 and M3 use the same inner-shadow measurements. M1 also uses ground-shadow measurements, while M3 uses estimated solar geometry.

\begin{figure*}[t]
\centering
\includegraphics[width=0.95\textwidth]{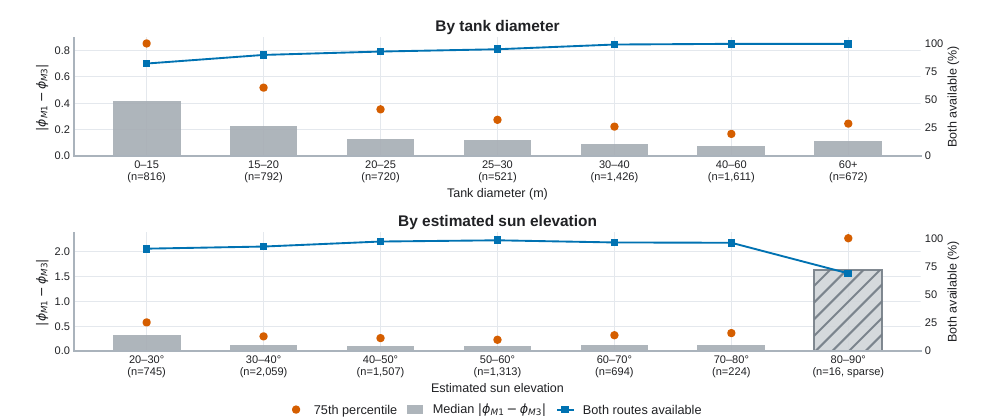}
\caption{NAIP internal-consistency diagnostics by annotated diameter (top) and estimated sun elevation (bottom). The figure labels M1 and M3 denote the displacement-ratio and affine estimators, respectively. Bars show their median absolute difference in normalized roof state \(\phi_R\); points show the 75th percentile; the blue line shows the share with both estimators available. Bin counts include all 6,558 tanks, with diameters of 60 m or more grouped together. Disagreement summaries condition on both estimates being available. The 80--90$^\circ$ sun bin is sparse (n=16), so its high disagreement is exploratory. NAIP has no ground-truth fill labels; the estimates assess consistency under inferred sun geometry, the nadir assumption, and the diameter-based height prior.}
\label{fig:supp_real_envelope}
\end{figure*}

The geometric and estimator-agreement checks require usable inner- and ground-shadow points, a constrained-versus-unconstrained direction change of at most 45 degrees, a solar-azimuth mismatch of at most 15 degrees, an inferred ground-shadow height from 6 to 30 m, and unclipped displacement-ratio and affine estimates within 0.15 of each other, with neither estimate below -0.05. Table~\ref{tab:supp_naip} gives pass rates by size; Fig.~\ref{fig:supp_real_envelope} shows agreement and estimate availability by estimated sun angle and tank diameter. Overall, 3,261/6,558 tanks (49.7\%) pass, including 150/336 (44.6\%) in held-out detector test chips. The affine estimator returns an estimate for 98\% of these tanks and the displacement-ratio estimator for 95\%. The median absolute difference between the two estimates is 0.049 among passing tanks and 0.104 among tanks with both estimates. The checks select partly on their agreement, so these rates measure internal consistency.

\begin{table}[t]
\centering
\caption{Share of NAIP tanks passing the geometric and estimator-agreement checks, by annotated diameter. There is no real fill ground truth.}
\label{tab:supp_naip}
\small
\begin{tabular}{@{}lrr@{}}
\toprule
Diameter & Tanks & Pass \\
\midrule
\(<15\) m & 816 & 12\% \\
15--25 m & 1,512 & 34\% \\
25--40 m & 1,947 & 58\% \\
\(\geq40\) m & 2,283 & 67\% \\
\bottomrule
\end{tabular}
\end{table}

\begin{figure*}[t]
\centering
\includegraphics[width=\textwidth]{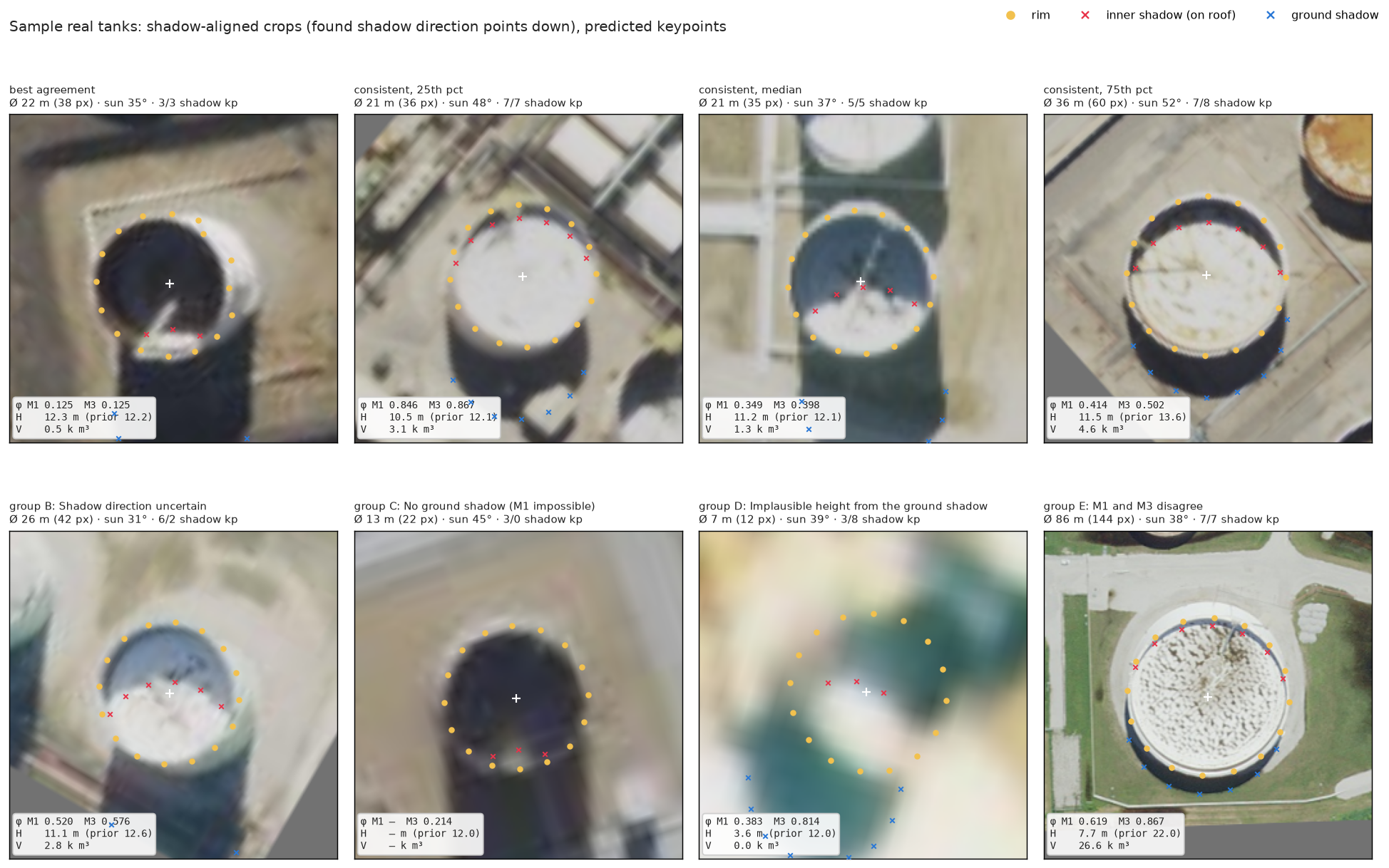}
\caption{Rule-selected real NAIP examples with predicted keypoints. Top row: tanks passing the geometric and estimator-agreement checks at selected M1--M3 disagreement quantiles. Bottom row, left to right: median-severity examples of uncertain shadow direction, missing ground shadow, implausible inferred height, and M1/M3 disagreement (outcome categories B--E). Image cards label the displacement-ratio and affine estimators M1 and M3, respectively, and report their normalized roof-state estimates, inferred height H (with its prior in parentheses), and model-derived volume V. These estimates use the stated assumptions; NAIP fill labels are unavailable.}
\label{fig:supp_naip_gallery}
\end{figure*}

Figure~\ref{fig:supp_naip_gallery} shows rule-selected application outcomes: four examples passing the checks at the 0th, 25th, 50th, and 75th percentiles of M1--M3 disagreement, followed by one median-severity example from each outcome category B--E. The gallery shows model outputs and outcome categories; NAIP ground-truth fill, height, and volume labels are unavailable.

We evaluate rotation against known sun directions on synthetic test images. On 1,249 synthetic test images, the sun-constrained search recovers azimuth within 10 degrees for 53.4\% and makes a greater-than-90-degree direction flip in 3.44\%. Restricting to sun elevations of 30--70 degrees (819 synthetic images), these rates are 62.0\% and 1.22\%. Without the date/location prior, the corresponding full-test rates are 48.7\% and 9.93\%, and the 30--70-degree rates are 55.1\% and 2.93\%. NAIP capture-time direction is unknown, so these results characterize the procedure on synthetic imagery.

\section{Shared-class detector transfer}
\label{supp:detection}

The transfer comparison uses the two classes present in both evaluation sets: external floating-roof tanks and closed-roof tanks. All three YOLO26s-OBB models are evaluated on the same 1,365 real test chips and 2,895 synthetic test tiles. The real set contains 527 floating-roof and 2,603 closed-roof objects; the synthetic set contains 14,347 and 2,458 objects, respectively. Synthetic-only training includes only these two classes, so we report their unweighted mean AP50--95 as the shared-class macro score.

The detector runs use 20 epochs, four GPUs, 1,024-pixel input, and the same augmentation setup. The real split removes 343 duplicate chips and assigns 1,429 groups of chips that share a USGS quarter-quadrangle, overlap, or are near-duplicates intact to train, validation and test (21,837 / 4,094 / 1,365 chips). Real-only training uses 21,837 unique chips, oversampled to 26,353 list entries; synthetic-only training uses 49,189 tiles from 19,990 synthetic training images. Mixed training combines these lists. Real-only and mixed checkpoints are selected on real validation AP50--95 (epochs 20 and 16), and the synthetic-only checkpoint on synthetic validation AP50--95 (epoch 20).

\begin{table*}[t]
\centering
\caption{Per-class and shared-class detector AP50--95 on identical test images. Shared-class mAP is the unweighted macro mean of floating-roof and closed-roof AP.}
\label{tab:supp_det}
\small
\begin{tabular}{@{}lrrrrrr@{}}
\toprule
Training data & Real floating & Real closed & Real shared mAP & Synthetic floating & Synthetic closed & Synthetic shared mAP \\
\midrule
Real only & 0.8924 & 0.8610 & 0.8767 & 0.8415 & 0.7255 & 0.7835 \\
Synthetic only & 0.6164 & 0.6000 & 0.6082 & 0.9756 & 0.9659 & 0.9708 \\
Mixed & 0.8970 & 0.8564 & 0.8767 & 0.9735 & 0.9638 & 0.9686 \\
\bottomrule
\end{tabular}
\end{table*}

Each detector condition has one training run, so the results describe these trained checkpoints. On these shared classes, synthetic-only training transfers to real imagery with a macro AP50--95 of 0.6082, while real-only training reaches 0.7835 on synthetic imagery. At confidence 0.25 and box IoU 0.5, the synthetic-only detector recovers 490 of the 527 real floating-roof test tanks with the correct class (93.0\%), against 470 (89.2\%) for the real-only detector. The synthetic-only detector produces 6,557 false positives on the real test chips, compared with 1,047 for the real-only detector. Of these, 5,743 occur on background objects absent from the synthetic images, including vehicles, rail cars, rooftop units and treatment basins. Mixed training matches real-only on the real shared-class macro score and approaches synthetic-only on synthetic imagery.

%
%

\bibliographystyle{ieeenat_fullname}
\bibliography{main}